\documentclass[runningheads]{llncs}
\usepackage{graphicx}
\usepackage{comment}
\usepackage{amsmath,amssymb} 
\usepackage{color}
\usepackage{hyperref}
\hypersetup{
    colorlinks=true,
    linkcolor=blue,
    filecolor=magenta,      
    urlcolor=blue,
}

\usepackage{multirow}
\usepackage{subcaption}
\usepackage{amsmath,amsfonts,bm}

\def\eqref#1{equation~\ref{#1}}

\def\1{\bm{1}}

\newcommand{\test}{\mathcal{D_{\mathrm{test}}}}

\def\vv{{\bm{v}}}

\def\mC{{\bm{C}}}

\def\mO{{\bm{O}}}

\def\mS{{\bm{S}}}

\DeclareMathAlphabet{\mathsfit}{\encodingdefault}{\sfdefault}{m}{sl}
\SetMathAlphabet{\mathsfit}{bold}{\encodingdefault}{\sfdefault}{bx}{n}
\newcommand{\tens}[1]{\bm{\mathsfit{#1}}}

\def\tD{{\tens{D}}}
\def\tE{{\tens{E}}}

\def\tG{{\tens{G}}}

\def\tI{{\tens{I}}}

\def\tL{{\tens{L}}}

\def\tW{{\tens{W}}}
\def\tX{{\tens{X}}}

\newcommand{\R}{\mathbb{R}}

\newcommand{\etal}{\textit{et al}.}

\begin{document}
\pagestyle{headings}
\mainmatter
\def\ECCVSubNumber{3098}  

\title{Context-Gated Convolution} 

\titlerunning{Context-Gated Convolution} 
\authorrunning{ECCV-20 submission ID \ECCVSubNumber} 

\author{Xudong Lin\inst{1}\thanks{This work was done when Xudong Lin interned at Tencent AI Lab.} 
\and
Lin Ma\inst{2}
\and
Wei Liu\inst{2}
\and
Shih-Fu Chang\inst{1}
}
\authorrunning{X. Lin et al.}
\institute{Columbia University \\
\email{\{xudong.lin, shih.fu.chang\}@columbia.edu} \and Tencent AI Lab \\
\email{forest.linma@gmail.com\qquad wl2223@columbia.edu}}


\maketitle

\begin{abstract}

As the basic building block of Convolutional Neural Networks (CNNs), the convolutional layer is designed to extract local patterns and lacks the ability to model global context in its nature. Many efforts have been recently devoted to complementing CNNs with the global modeling ability, especially by a family of works on global feature interaction. In these works, the global context information is incorporated into local features before they are fed into convolutional layers. However, research on neuroscience reveals that the neurons' ability of modifying their functions dynamically  according to context is essential for the perceptual tasks, which has been overlooked in most of CNNs. Motivated by this, we propose one novel Context-Gated Convolution (CGC) to explicitly modify the weights of convolutional layers adaptively under the guidance of global context. As such, being aware of the global context, the modulated convolution kernel of our proposed CGC can better extract representative local patterns and compose discriminative features. Moreover, our proposed CGC is lightweight and applicable with modern CNN architectures, and consistently improves the performance of CNNs according to extensive experiments on image classification, action recognition, and machine translation. Our code of this paper is available at \url{https://github.com/XudongLinthu/context-gated-convolution}.

\end{abstract}

\section{Introduction}

\begin{figure}
    \centering
    \includegraphics[width=.99\textwidth]{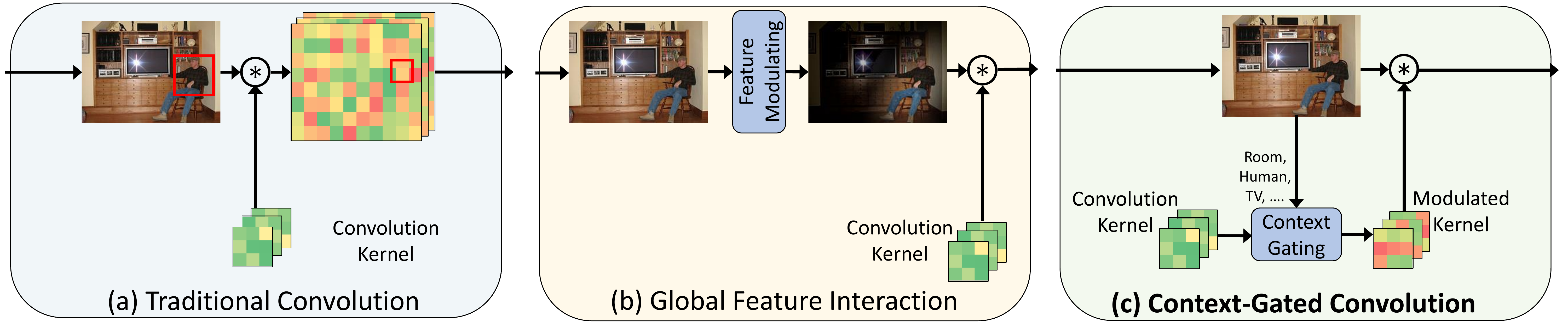}
    \caption{(a) Traditional convolution only composes local information. (b) Global feature interaction methods modify input feature maps by incorporating global information. (c) Our proposed CGC, in a fundamentally different manner, modulates convolution kernels under the guidance of global context. $\circledast$ denotes convolution.}\label{idea}
\end{figure}

Convolutional Neural Networks (CNNs) have achieved remarkable successes on various tasks, e.g., image classification~\cite{He_2016_CVPR,huang2017densely}, object detection~\cite{rcnn,fasterrcnn}, image translation~\cite{CycleGAN2017}, action recognition~\cite{Kinetics}, sentence/text classification~\cite{zhang2015character,kim2014convolutional}, machine translation~\cite{gehring2017convolutional}, etc. However, the sliding window mechanism of convolution makes it only capable of capturing local patterns, limiting its ability of utilizing global context. Taking the 2D convolution on the image as one example, as Fig.~\ref{idea}(a) shows, the traditional convolution only operates on the local image patch and thereby composes local features.

According to the recent research on neuroscience~\cite{li2004perceptual,gilbert2013top}, neurons' awareness of global context is important for us to better interpret visual scenes, stably perceive objects, and effectively process complex perceptual tasks. Many methods~\cite{vaswani2017attention,wang2017residual,wang2017non,park2018bam,hu2018squeeze,chen2019graph,cao2019gcnet,bello2019attention,li2019selective} have been recently proposed to introduce global context modeling modules into CNN architectures. As Fig.~\ref{idea}(b) shows, these methods, which are named as global feature interaction methods in this paper, modulate intermediate feature maps by incorporating the global context into the local feature representation.

However, as stated in~\cite{gilbert2013top}, ``\textit{rather than having a fixed functional role, neurons should be thought of as adaptive processors, changing their function according to the behavioural context}". Therefore, the context information should be utilized to explicitly modulate the convolution kernels for ``\textit{changing the structure of correlations over neuronal ensembles}''~\cite{gilbert2013top}. However, to the best of our knowledge, such a modulating mechanism has not been exploited in CNNs yet, even though it is one efficient and intuitive way. 
Motivated by this, we will model convolutional layers as ``\textit{adaptive processors}'' and explore how to leverage global context to guide the composition of local features in convolution operations.

In this paper, we propose Context-Gated Convolution (CGC), as shown in Fig.~\ref{idea}(c), a new perspective of complementing CNNs with the awareness of the global context. Specifically, our proposed CGC learns a series of mappings to generate gates from the global context feature representations to modulate convolution kernels accordingly. With the modulated kernels, the traditional convolution is performed on input feature maps, which enables convolutional layers to dynamically capture representative local patterns and compose local features of interest under the guidance of global context. Our contributions lie in three-fold.
\begin{itemize}
    \item To the best of our knowledge, we make the first attempt of introducing the context-awareness to convolutional layers by modulating their weights according to the global context.
    
    \item We propose a novel lightweight Context-Gated Convolution (CGC) to effectively generate gates for convolution kernels to modify the weights with the guidance of global context. Our CGC consists of a Context Encoding Module that encodes context information into latent representations, a Channel Interacting Module that projects them into the space of output dimension, and a Gate Decoding Module that decodes the latent representations to produce the gate.
    
    \item Our proposed CGC can better capture local patterns and compose discriminative features, and consistently improve the generalization of traditional convolution with a negligible complexity increment on various tasks including image classification, action recognition, and machine translation.
\end{itemize}

\section{Related Works}
There have been many efforts in augmenting CNNs with context information. They can be roughly categorized into three types: first, adding backward connections in CNNs~\cite{stollenga2014deep,zamir2017feedback,yang2018convolutional} to model the top-down influence~\cite{gilbert2013top} like humans' visual processing system; second, modifying intermediate feature representations in CNNs according to the attention mechanism~\cite{vaswani2017attention,wang2017non,woo2018cbam,chen2019graph,cao2019gcnet}; third, dynamically generating the parameters of convolutional layers according to local or global information~\cite{jia2016dynamic,dai2017deformable,wu2019pay,zhu2019deformable,jo2018deep,mildenhall2018burst,cheng2018dual}.

For the first category of works, it is still unclear how the feedback mechanism can be effectively and efficiently modeled in CNNs. For example, Yang \etal~\cite{yang2018convolutional} proposed an Alternately Updated Clique to introduce feedback mechanisms into CNNs. However, compared to traditional CNNs, the complex updating strategy increases the difficulty for training them as well as the latency at the inference time. The second category of works is the global feature interaction methods. They \cite{vaswani2017attention,wang2017residual,wang2017non,park2018bam,hu2018squeeze,woo2018cbam,chen2019graph,cao2019gcnet,bello2019attention} were proposed recently to modify local features according to global context information, usually by a global correspondence, i.e., the self-attention mechanism. There are also works on reducing the complexity of the self-attention mechanism~\cite{parmar2018image,child2019generating}. However, this family of works only considers changing the input feature maps. 

The third type of works is more related to our work. Zhu \etal~\cite{zhu2019deformable} proposed to adaptively set the offset of each element in a convolution kernel and the gate value for each element in the input local feature patch. However, the mechanism only changes the input to the convolutional layer. The weight tensor of the convolutional layer is not considered. Wu \etal~\cite{wu2019pay} proposed to dynamically generate the weights of convolution kernels. However, it is specialized for Lightweight Convolution~\cite{wu2019pay} and only takes local segments as inputs. Another family of works on dynamic filters~\cite{jia2016dynamic,jo2018deep,mildenhall2018burst} also belongs to this type. They generate weights of convolution kernels using features extracted from input images by another CNN feature extractor. The expensive feature extraction process makes it more suitable for generating a few filters, e.g., in the case of low-level image processing. It is impractical to generate weights for all the layers in a deep CNN model in this manner.

\section{Context-Gated Convolution}

\subsection{Preliminaries}
Without loss of generality, we consider one sample of 2D case. The input to a convolutional layer is a feature map $\tX \in \R^{c \times h \times w}$, where $c$ is the number of channels, and $h,w$ are respectively the height and width of the feature map. In each convolution operation, a local patch of size $c \times k_1 \times k_2$ is collected by the sliding window to multiply with the kernel $\tW \in \R^{o \times c \times k_1 \times k_2}$ of this convolutional layer, where $o$ is the number of output channels, and $k_1,k_2$ are respectively the height and width of the kernel. Therefore, only local information within each patch is extracted in one convolution operation. Although in the training process, the convolution kernels are learned from all the patches of all the images in the training set, the kernels are not adaptive to the current context during the inference time.

\begin{figure*}[t]
\begin{center}

\includegraphics[width=0.8\linewidth]{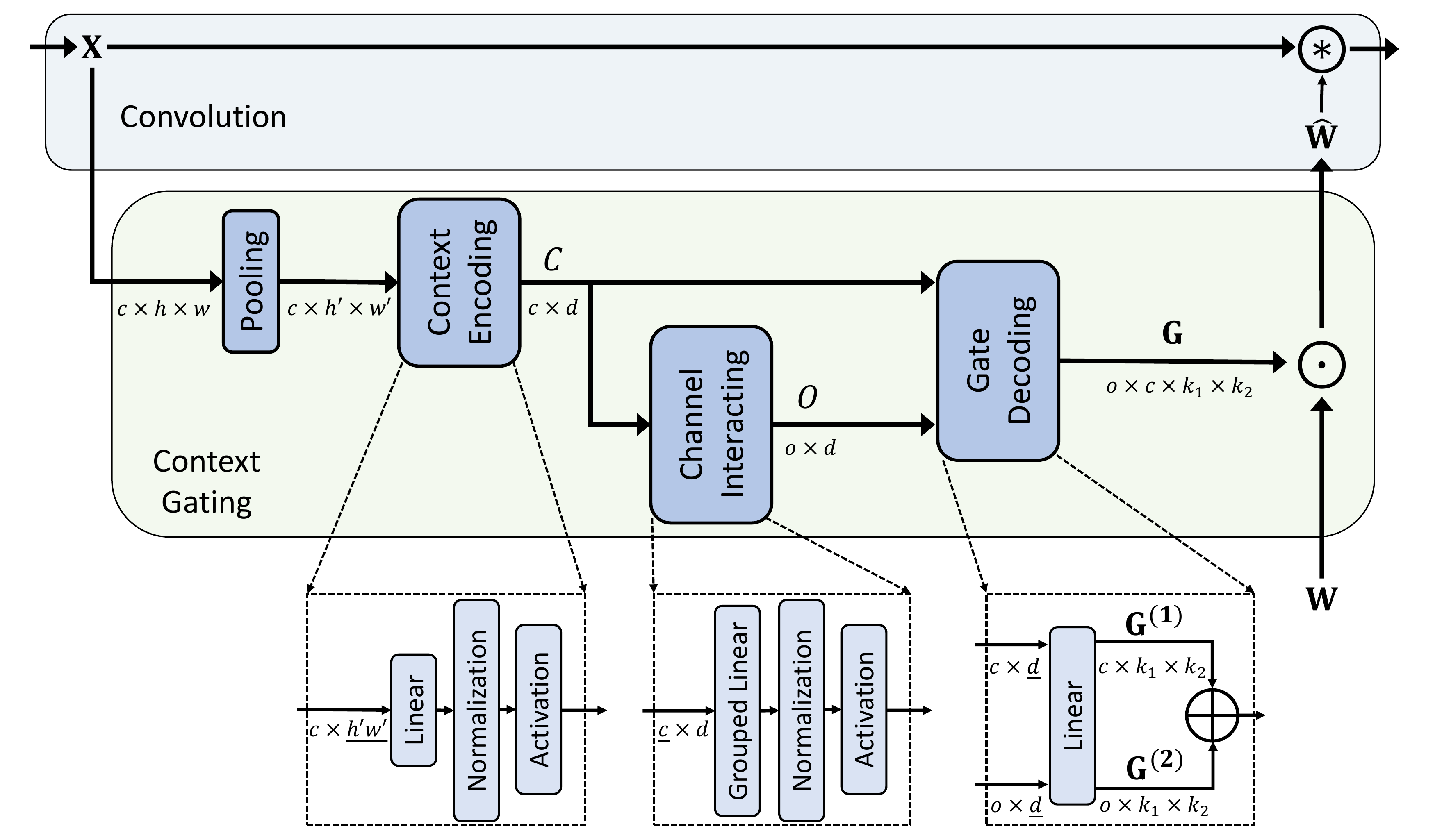}
\end{center}
\caption{\label{fig:framework}Our proposed CGC consists of three components, namely the Context Encoding Module, the Channel Interacting Module, and the Gate Decoding Module. The Context Encoding Module encodes global context information into a latent representation $\mC$; the Channel Interacting Module transforms $\mC$ to $\mO$ with output dimension $o$; the Gate Decoding Module produces $\tG^{(1)}$ and $\tG^{(2)}$ from $\mC$ and $\mO$ to construct the gate $\tG$. $\circledast$ and $\odot$ denote convolution and element-wise multiplication operations, respectively. $\oplus$ is shown in Eq.~(\ref{sum}). The dimension transformed in each linear layer is underlined.}
\end{figure*}

\subsection{Module Design}
In order to handle the aforementioned drawback of traditional convolution, we propose to incorporate the global context information during the convolution process. Different from the existing approaches that modify the input features according to the context, e.g., a global correspondence of feature representations, we attempt to directly modulate the convolution kernel under the guidance of the global context information.

One simple and straightforward way of modulating the convolution kernel $\tW $ with global context information is to directly generate a gate $\tG \in \R^{o \times c \times k_1 \times k_2}$ of the same size as $\tW$ according to the global context. Assuming that we generate the gate from a context vector $\vv \in \R^l$ using a linear layer without the bias term, the number of parameters is $l \times o \times c \times k_1 \times k_2$, which is extremely catastrophic when we modulate the convolution kernel of every convolutional layer. For modern CNNs, $o$ and $c$ can be easily greater than 100 or even 1,000, which makes $o\times c$ the dominant term in the complexity. Inspired by previous works on convolution kernel decomposition~\cite{howard2017mobilenets,chollet2017xception}, we propose to decompose the gate $\tG$ into two tensors $\tG^{(1)} \in \R^{ c \times k_1 \times k_2}$ and $\tG^{(2)} \in \R^{o  \times k_1 \times k_2}$, so that the complexity of $o \times c$ can thereby significantly be broken down.

However, directly generating these two tensors is still impractical. Supposing that we generate them with two linear layers, the number of parameters is $l \times (o + c) \times k_1 \times k_2$, which is of the same scale as the number of parameters of the convolution kernel itself. The bottleneck now is jointly modeling channel-wise and spatial interactions, namely $l$ and $(o+c) \times k_1 \times k_2$, considering that $\vv \in \R^l$ is encoded from the input feature map $\tX \in \R^{c \times h \times w}$. Inspired by depth-wise separable convolutions~\cite{howard2017mobilenets,chollet2017xception}, we propose to model the spatial interaction and the channel-wise interaction separately to further reduce the complexity. 

In this paper, we propose one novel Context-Gated Convolution (CGC) to incorporate the global context information during the convolution process. Specifically, our proposed CGC consists of three modules: the Context Encoding Module, the Channel Interacting Module, and the Gate Decoding Module. As shown in Fig.~\ref{fig:framework}, the Context Encoding Module encodes global context information in each channel into a latent representation $\mC$ via spatial interaction; the Channel Interacting Module projects the latent representation to the space of output dimension $o$ via channel-wise interaction; the Gate Decoding Module produces $\tG^{(1)}$ and $\tG^{(2)}$ from the latent representation $\mC$ and the projected representation $\mO$ to construct the gate $\tG$ via spatial interaction. The detailed information is described in the following.

\textbf{Context Encoding Module.}
To extract contextual information, we first use a pooling layer to reduce the spatial resolution to $h' \times w'$ and then feed the resized feature map to the Context Encoding Module. It encodes information from all the spatial positions for each channel, and extracts a latent representation of the global context. We use a linear layer with weight $\tE \in \R^{h' \times w' \times d}$ to project the resized feature map in each channel to a latent vector of size $d$. Inspired by the bottleneck structure from~\cite{He_2016_CVPR,hu2018squeeze,wang2017non,vaswani2017attention}, we set $d=\frac{k_1 \times k_2}{2}$ to extract informative context, when not specified. The weight $\tE$ is shared across different channels. A normalization layer and an activation function come after the linear layer. There are $c$ channels, so the output of the Context Encoding Module is $\mC \in \R^{c \times d}$. Since the output is fed into two different modules, we accordingly apply two individual Normalization layers so that different information can be conveyed if needed.

\textbf{Channel Interacting Module.}
It projects the feature representations $C \in \R^{c \times d}$ to the space of output dimension $o$. Inspired by \cite{ha2016hypernetworks}, we use a grouped linear layer $\tI \in \R^{\frac{c}{g} \times \frac{o}{g}}$, where $g$ is the number of groups. The weight $\tI$ is shared among different dimensions of $d$ and different groups. A normalization layer and an activation function come after the linear layer. The final output of the Channel Interacting Module is $\mO \in \R^{o \times d}$.

\textbf{Gate Decoding Module.}
It takes both $\mC$ and $\mO$ as inputs, and decodes the latent representations to the spatial size of convolution kernels. We use two linear layers whose weights $\tD_c \in \R^{d \times k_1 \times k_2}$ and $\tD_o \in \R^{d \times k_1 \times k_2}$ are respectively shared across different channels in $\mC$ and $\mO$. Then each element in the gate $\tG$ is produced by:
\begin{equation}
\label{sum}
    \tG_{h,i,j,k} = \sigma(\tG^{(1)}_{i,j,k}+\tG^{(2)}_{h,j,k})=\sigma((\mC \tD_c)_{i,j,k} + (\mO \tD_o)_{h,j,k}),
\end{equation}
where $\sigma(\cdot)$ denotes the sigmoid function. Now we have $\tG$ with the same size of the convolution kernel $\tW$, which is generated from the global context by our lightweight modules. Then we can modulate the weight of a convolutional layer by element-wise multiplication to incorporate rich context information:
\begin{equation}
    \hat{\tW} = \tW \odot \tG.
\end{equation}
With the modulated kernel, a traditional convolution process is performed on the input feature maps, where the context information can help the kernel capture more representative patterns and also compose features of interest.

\textbf{Complexity.}
The computational complexity of our three modules is $O(c \times d \times h' \times w'+c \times o /g+c \times d \times k_1 \times k_2+o \times d \times k_1 \times k_2+o \times c \times k_1 \times k_2)$, where $h',w'$ can be set independent of $h,w$. It is negligible compared to convolution's $O(o \times c \times k_1\times k_2 \times h \times w)$. Except the linear time of pooling, the complexity of these three modules is independent of the input's spatial size. The total number of parameters is $O(d \times h' \times w'+c \times o /g^2+d \times k_1 \times k_2)$, which is negligible compared to traditional convolution's $O(o \times c \times k_1 \times k_2)$. Therefore, we can easily replace the traditional convolution with our proposed CGC with a very limited computation and parameter increment, and enable convolutional layers to be adaptive to global context.

\subsection{Discussions}
We are aware of the previous works on dynamically modifying the convolution operation~\cite{dai2017deformable,wu2019pay,jia2016dynamic,jo2018deep,mildenhall2018burst,zhu2019deformable}. As discussed before, \cite{zhu2019deformable} essentially changes the input to the convolutional layer but not the weight tensor of the convolutional layer. Dynamic Convolution~\cite{wu2019pay} is specialized for the Lightweight convolution~\cite{wu2019pay} and only adaptive to local inputs. The family of work on dynamic filters~\cite{jia2016dynamic,jo2018deep,mildenhall2018burst} generates weights of convolution kernels using features extracted from input images by another CNN feature extractor. It is too expensive to generate weights for all the layers in a deep CNN model in this manner. In contrast, our CGC takes feature maps of a convolutional layer as input and makes it possible to dynamically modulate the weight of each convolutional layer, which systematically improves CNNs' global context modeling ability.

Both global feature interaction methods\cite{vaswani2017attention,wang2017residual,wang2017non,park2018bam,hu2018squeeze,chen2019graph,cao2019gcnet,bello2019attention} modifying feature maps~ and our proposed CGC modulating kernels can incorporate the global context information into CNN architectures, which can boost the performance of CNNs. However, 1) with our CGC, the complexity of modulating kernels does not depend on input size, but global feature interaction methods, e.g., Non-local, may suffer from a quadratic computational complexity w.r.t. the input size; 2) our CGC can be easily trained from scratch and improve the training stability of CNNs according to our experiments (Sections~\ref{sec:image} and ~\ref{sec:act}); 3) by modulating kernels, our CGC can dynamically create kernels with specialized functions according to context (Section~\ref{sec:image}) and thus enable CNNs to accordingly capture discriminative information as adaptive processors, which cannot be realized by modifying feature maps. Moreover, our CGC is also somewhat complementary to global feature interaction methods (Sections~\ref{sec:image} and ~\ref{sec:act}) and we can further improve CNN's performance by applying both CGC and global feature interaction methods.

\section{Experiments}
In this section, we demonstrate the effectiveness of our proposed CGC in incorporating 1D, 2D, and 3D context information in 1D, 2D, and (2+1)D convolutions. We conduct extensive experiments on image classification, action recognition, and machine translation, and observe that our CGC consistently improves the performance of modern CNNs with a negligible parameter increment on six benchmark datasets: ImageNet~\cite{russakovsky2015imagenet}, CIFAR-10~\cite{krizhevsky2009learning}, ObjectNet~\cite{barbu2019objectnet}, Something-Something (v1)~\cite{Goyal_2017_ICCV}, Kinetics~\cite{Kinetics}, and IWSLT'14 De-En~\cite{cettolo2014report}.

\subsection{Implementation Details}

All of the experiments are based on PyTorch~\cite{pytorch}. All the linear layers are without bias terms. We follow common practice to use Batch Normalization~\cite{ioffe2015batch} for computer vision tasks, and Layer Normalization~\cite{ba2016layer} for natural language processing tasks, respectively. We use ReLU~\cite{nair2010rectified} as the activation function for all the experiments in this paper. We use average pooling with $h'=k_1$ and $w'=k_2$, when not specified. Note that we only replace the convolution kernels with a spatial size larger than $1$. For those point-wise convolutions, we take them as linear layers and do not modulate them. To reduce the size of $\tI$, we fix $c/g=16$ when not specified. We initialize all these layers as what \cite{He_2015} did for computer vision tasks and as what \cite{pmlr-v9-glorot10a} did for natural language processing tasks, when not specified.

\begin{table*}[t]
\caption{Image classification results on ImageNet and CIFAR-10. Param indicates the number of parameters in the model. $\Delta$MFLOPs is the increment of the number of multiplication-addition operations compared to ResNet-50 (R50, 4 GFLOPs) for ImageNet models and ResNet-110 (R110, 256 MFLOPs) for CIFAR-10 models. Bold indicates the best result.} 
\label{tab:img}
\tiny
\begin{center}
\begin{tabular}{lllllll}
\hline
Dataset & Training Setting & Model  & Param & $\Delta$MFLOPs & Top-1(\%) & Top-5(\%) 
\\ \hline 
\multirow{18}*{ImageNet} & \multirow{3}*{-} & R50 + GloRe~\cite{chen2019graph} &  30.5M & 1200 & 78.4 & -\\
~&~&DCNv2-R50~\cite{zhu2019deformable} & 27.4M & 200 & 78.2 & 94.0 \\
~&~&GC-R50~\cite{cao2019gcnet} & 28.08M & 100 & 77.70 & 93.66 \\
\cline{2-7}
~ & \multirow{9}*{Default} & SE-R50~\cite{hu2018squeeze} & 28.09M & 8 & 77.18 & 93.67 \\
~&~&BAM-R50~\cite{park2018bam} & 25.92M & 83 & 76.90 & 93.40 \\
~&~&GC-R50~\cite{cao2019gcnet} & 28.11M & 8 & 73.90 & 91.70 \\ 
~&~&DCNv2-R50~\cite{zhu2019deformable} & 27.4M & 200 & 77.21 & 93.69 \\
~&~&SK-R50~\cite{li2019selective} & 37.25M & 1837 & 77.15 & 93.54 \\
\cline{3-7}

~&~&R50~\cite{He_2016_CVPR}         & 25.56M & - & 76.16 & 92.91 \\ 
~&~&\textbf{R50 + CGC(Ours)} & \textbf{25.59M} & \textbf{6} & \textbf{77.48} & \textbf{93.81} \\
\cline{3-7}
~&~&CBAM-R50~\cite{woo2018cbam} & 28.09M & 15 & 77.34 & \textbf{93.69} \\
~&~&\textbf{CBAM-R50 + CGC(Ours)} & 28.12M & 21 & \textbf{77.68} & 93.68 \\
\cline{2-7}

~&\multirow{6}*{Advanced}&DCNv2-R50~\cite{zhu2019deformable} & 27.4M & 200 & 78.89 & 94.60 \\
~&~& SE-R50~\cite{hu2018squeeze} & 28.09M & 8 & 78.79 & 94.52 \\
\cline{3-7}
~&~&R0~\cite{He_2016_CVPR}         & 25.56M & - & 78.13 & 94.06 \\ 
~&~&\textbf{R50 + CGC(Ours)} & \textbf{25.59M} & \textbf{6} & \textbf{79.54} & \textbf{94.78} \\
\cline{3-7}
~&~&CBAM-R50~\cite{woo2018cbam} & 28.09M & 15 & 78.86 & 94.58 \\
~&~&\textbf{CBAM-R50 + CGC(Ours)} & 28.12M & 21 & \textbf{79.74} & \textbf{94.83} \\
\hline
\multirow{2}*{CIFAR-10} &~& R110~\cite{he2016identity}         & 1.73M & - & 93.96 & 99.73 \\ 
~&~&\textbf{R110 + CGC(Ours)}        & \textbf{1.80M} & \textbf{2} & \textbf{94.86} & \textbf{99.82} \\ 
\hline
\end{tabular}
\end{center}

\end{table*}

\subsection{Image Classification}
\label{sec:image}
\textbf{Experimental Setting.} Following previous works~\cite{He_2016_CVPR} on ImageNet~\cite{russakovsky2015imagenet}, we train models on the ImageNet 2012 training set, which contains about 1.28 million images from 1,000 categories, and report the results on its validation set, which contains 50,000 images. We replace all the convolutions that are not $1\times 1$ in ResNet-50~\cite{He_2016_CVPR} with our CGC and train the network from scratch. Note that for the first convolutional layer, we use $\tI \in \R^{3\times 64}$ for the Channel Interacting Module. We conduct experiments in two settings: Default and Advanced. For the Default setting, we follow common practice~\cite{He_2016_CVPR} and apply minimum training tricks. For the Advanced setting, we borrow training tricks from \cite{he2019bag} to validate that our CGC can still improve the performance, even under a strong baseline. CIFAR-10 contains 50K training images and 10K testing images in 10 classes. We follow common practice~\cite{he2016identity} to train and evaluate the models. We take ResNet-110~\cite{he2016identity} (with plain blocks) as the baseline model. All the compared methods are trained based on the same training protocol. The details are provided in the supplementary material. For evaluation, we report Top-1 and Top-5 accuracies of a single crop with the size $224\times 224$ for ImageNet and $32\times 32$ for CIFAR-10, respectively.

ObjectNet~\cite{barbu2019objectnet} is a new challenging evaluation dataset for image classification. There are 113 classes out of 313 ObjectNet classes, which overlap with ImageNet classes. We follow~\cite{barbu2019objectnet} to evaluate models trained on ImageNet on the overlapped classes.

\textbf{Performance Results.}
As Table~\ref{tab:img} shows, our CGC significantly improves the performances of baseline models on both ImageNet and CIFAR-10. On ImageNet, our CGC improves the Top-1 accuracy of ResNet-50 under the Advanced setting by 1.41\% with only 0.03M more parameters and 6M more FLOPs, which verifies our CGC's effectiveness of incorporating global context and its efficiency. We observe that our CGC outperforms DCN-v2~\cite{zhu2019deformable}, SK-Net~\cite{li2019selective} and CBAM~\cite{woo2018cbam}, which indicates the superiority of modulating kernels. We also observe that our CGC can also improve the performance of CBAM-ResNet-50~\cite{woo2018cbam} consistently under both settings, which indicates that our proposed CGC is applicable with state-of-the-art global feature interaction methods. CBAM-ResNet-50 + CGC even reaches 79.74\% Top-1 accuracy, which outperforms the other compared methods by a large margin.

\begin{figure*}[t]
\begin{center}
\includegraphics[width=0.95\linewidth]{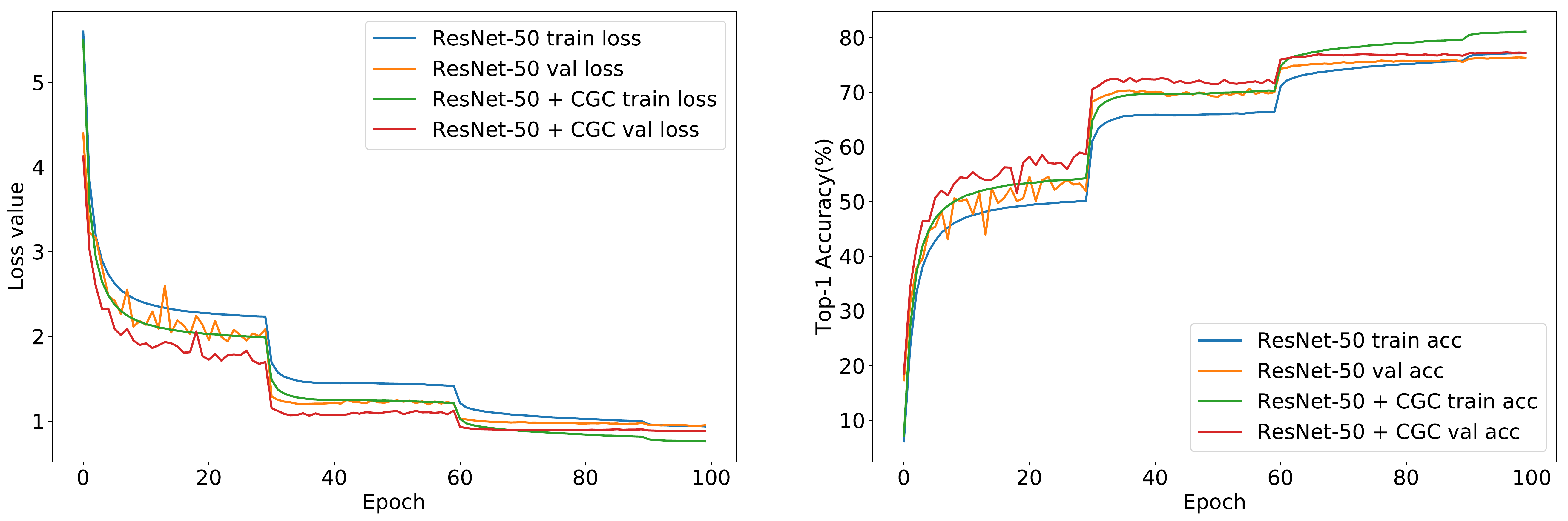}
\end{center}
\caption{\label{train}The training curves of ResNet-50 and ResNet-50 + CGC (ours) on ImageNet under the default training setting.}
\end{figure*}

We also find that GC-ResNet-50 is hard to train from scratch unless using the fine-tuning protocol reported by~\cite{cao2019gcnet}, which indicates that modifying features may be misleading in the early training process. Although our CGC introduces a few new parameters, our model converges faster and more stably compared to vanilla ResNet-50, as shown in Fig.~\ref{train}. We conjecture that this is because the adaptiveness to global context improves the model's generalization ability and the gating mechanism reduces the norm of gradients back-propagated to the convolution kernels, which leads to a smaller Lipschitz constant and thus better training stability~\cite{santurkar2018does,qiao2019weight}.

\begin{table*}[t]
\caption{Image classification results on ObjectNet. Bold indicates the best result.} 
\label{tab:obj}
\scriptsize
\begin{center}
\begin{tabular}{lll}
\hline
  Model   & Top-1(\%) & Top-5(\%) \\ \hline
  R50~\cite{He_2016_CVPR}         & 29.35 & 48.42 \\
  SE-R50~\cite{hu2018squeeze} & 29.48 & 45.55 \\
  DCNv2-R50~\cite{zhu2019deformable} & 29.74 & 48.83 \\
  CBAM-R50~\cite{woo2018cbam} & 29.56 & 48.68 \\
  \textbf{R50 + CGC(Ours)} & \textbf{31.53} & \textbf{50.16}
  
\\ \hline 

\hline
\end{tabular}
\end{center}

\end{table*}

To further validate the generalization ability of our CGC, we use ObjectNet to evaluate models with good performances on ImageNet. ObjectNet~\cite{barbu2019objectnet} is recently proposed to push image recognition models beyond their current limit of generalization. The dataset contains images ``in many rotations, on different backgrounds, from multiple viewpoints'', which makes it hard for models trained on ImageNet to correctly classify these images. As Table~\ref{tab:obj} shows, our CGC significantly improves the generalization ability of the ResNet-50 baseline. The improvement (2.18\%) is even larger than that on ImageNet validation set.

\begin{figure*}[t]

\begin{center}
\includegraphics[width=0.95\linewidth]{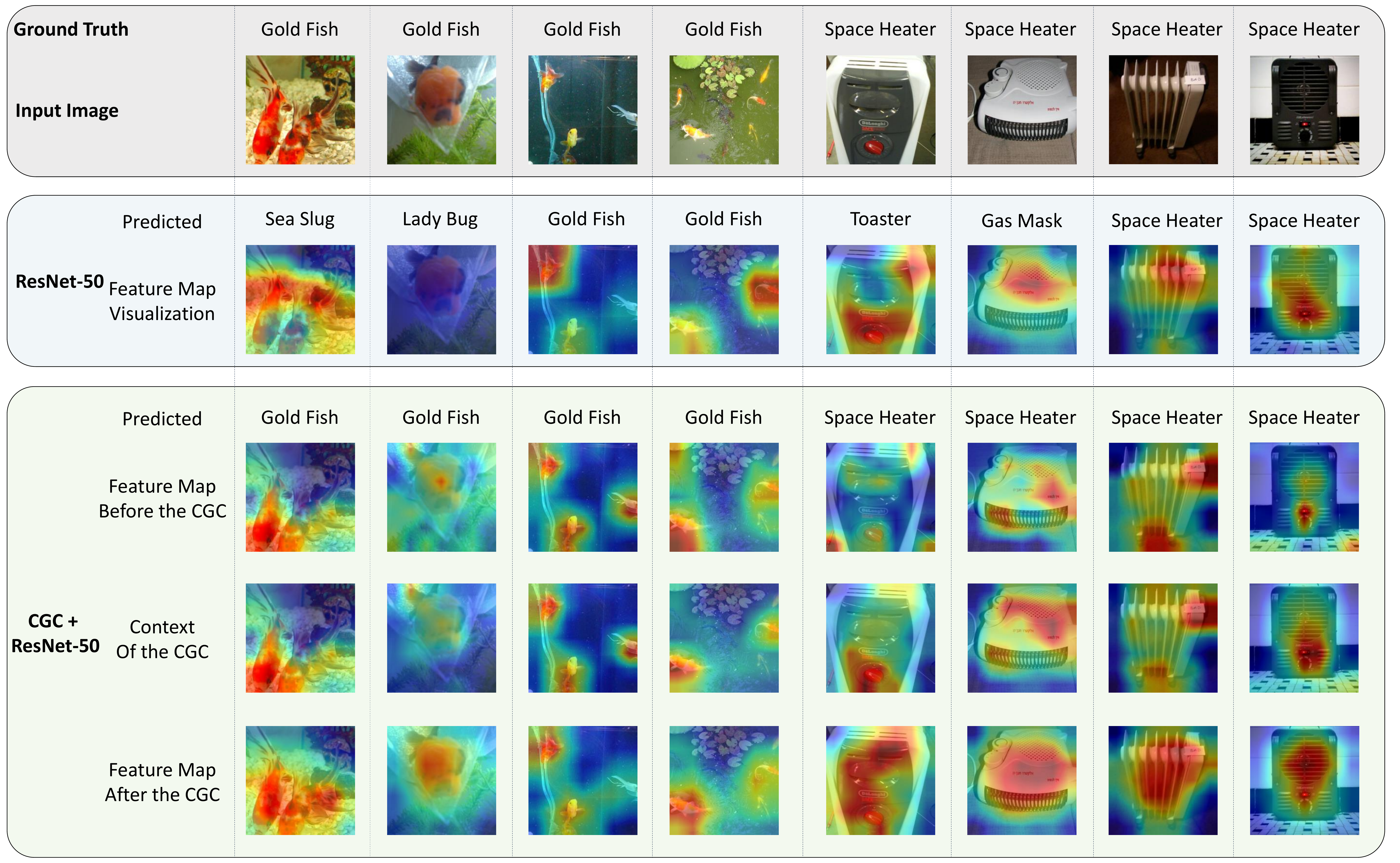}
\end{center}
\caption{\label{visual}Visualization of the feature maps produced by ResNet-50 and CGC-ResNet-50 from the ImageNet validation set images. (Best viewed on a monitor when zoomed in.)}
\end{figure*}

\textbf{Visualization.}
To understand how CGC helps the model capture more informative features under the guidance of context information, we visualize the feature maps of ResNet-50 and our CGC-ResNet-50 by Grad-CAM++~\cite{chattopadhay2018grad}. As Fig.~\ref{visual} shows, overall, the feature maps (After the CGC) produced by our CGC-ResNet-50 cover more informative regions, e.g., more instances or more parts of the ground-truth object, than vanilla ResNet-50.

Specifically, we visualize the feature maps before the last CGC in the model, the context information used by CGC, and the resulting feature maps after the CGC. As is clearly shown in Fig.~\ref{visual}, the proposed CGC extracts the context information from representative regions of the target object and successfully refines the feature maps with comprehensive understanding of the whole image and the target object. For example, in Gold Fish 1, the heads of the fishes are partially visible. Vanilla ResNet-50 mistakes this image as Sea Slug, because it only pays attention to the tails of the fishes, which are similar to sea slugs. However, our CGC utilizes the context of the whole image and guides the convolution with information from the entire fishes, which helps the model classify this image correctly.

\textbf{Analysis of the Gate.}
To further validate that our CGC uses context information of the target objects to guide the convolution process, we calculate the average modulated kernel (in the last CGC of the model) for images of each class in the validation set. Then we calculate inter-class $L-2$ distance between every two average modulated kernels, i.e., class centers, and the intra-class $L-2$ distance (mean distance to the class center) for each class. As is shown in the supplementary material, we visualize the difference matrix between inter-class distances and intra-class distances. In more than 93.99\% of the cases, the inter-class distance is larger than the corresponding intra-class distance, which indicates that there are clear clusters of these modulated kernels and the clusters are aligned very well with the classes.

This observation strongly supports that our CGC successfully extracts class-specific context information and effectively modulates the convolution kernel to extract representative features. Meanwhile, the intra-class variance of the modulated kernels indicates that our CGC dynamically modulates convolution kernels according to different input contexts.

\textbf{Ablation Study.} 
In order to demonstrate the effectiveness of our module design, ablation studies are conducted on CIFAR-10, as illustrated in Table \ref{tab:cifar}. Specifically, we ablate many variants of our CGC and find that our default setting is a good trade-off between parameter increment and performance gain. The experiments on the combination of $\tG^{(1)}$ and $\tG^{(2)}$ show that our decomposition approach in Eq.~(\ref{sum}) is a better way to construct the gate. For channel interacting, we find that using a full linear model with $g=1$ achieves better performance with more parameters, as is expected. We try removing the bottleneck structure and set $d=k_1\times k_2$, and the performance drops, which validates the necessity of the bottleneck structure.

Shared Norm indicates using the same Normalization layer for the following two branches. For Two $\tE$s, we learn another $\tE$ to encode $\mC$ only for the Channel Interacting Module.  We also try sharing $\tD$ for generating $\tG^{(1)}$ and $\tG^{(2)}$, using larger resized feature maps and using max pooling instead of average pooling. All the results support our default setting. We also test different numbers of layers to replace traditional convolutions with our CGC. The result indicates that the more, the better. We select 3 variants with a similar number of parameters and performance on CIFAR-10 and further perform ablation studies for them on ImageNet. As Table~\ref{tab:abimg} shows, we observe the same performance rankings of these variants on ImageNet as those on CIFAR-10.

\begin{table}[t] 
\caption{Ablation studies on CIFAR-10 and ImageNet. Param denotes the number of parameters in the model. $\Delta$MFLOPs is the increment of the number of multiplication-addition operations compared to ResNet-110 (256 MFLOPs). Bold indicates our default setting. Top-1 Accuracy (\%) (average of 3 runs) is reported.}

  \begin{minipage}[b]{0.5\textwidth} 
    \tiny
    \centering
    \subcaption{CIFAR-10}
    \label{tab:cifar}
    \begin{tabular}{llll}
    \hline
    Model  & Param & $\Delta$MFLOPs & Top-1 (\%) 
    \\ \hline 
    ResNet-110~\cite{he2016identity}         & 1.73M & - & 93.96  \\ 
    \textbf{ResNet-110 + CGC}        & 1.80M & 1.681 & 94.86  \\ 
    \hline
    only $\tG^{(1)}$         & 1.75M & 1.447 & 94.53  \\ 
    only $\tG^{(2)}$         & 1.78M & 1.472 & 94.41  \\ 
    $\tG^{(1)}*\tG^{(2)}$         & 1.80M & 1.681 & 94.59  \\ 
    \hline
    $g=1$              & 1.96M & 1.681 & 94.97  \\
    $d=k_1\times k_2$      & 1.81M & 1.741 & 94.61  \\ 
    \hline
    Shared Norm         & 1.79M & 1.681 & 94.72 \\
    Two $\tE$s              & 1.80M & 1.871 & 94.53  \\
    Shared $\tD$         & 1.79M & 1.681 & 94.78  \\
    \hline
    $h'=2k_1,w'=2k_2$              & 1.81M & 1.681 & 94.80  \\
     MaxPool              & 1.80M & 1.681 & 94.44  \\
    \hline
    (res1,2,3)         & 1.80M & 1.678 & 94.55  \\
    (res2,3)         & 1.78M & 1.052 & 94.43  \\
    (res3)         & 1.76M & 0.622 & 94.26  \\
    \hline
    
    \end{tabular} 
  \end{minipage}%
  \begin{minipage}[b]{0.5\textwidth} 
    \tiny
    \centering
    \subcaption{ImageNet}
    \label{tab:abimg}
    \begin{tabular}{llll}
    \hline
    Model   & Top-1 (\%)
    \\ \hline 
    ResNet-50~\cite{He_2016_CVPR}          & 76.16  \\ 
    \textbf{ResNet-50 + CGC}         & 77.48  \\ 
   
    \hline
    Shared Norm          & 77.21 \\
    Shared $\tD$        & 77.28  \\
    \hline
    $h'=2k_1,w'=2k_2$               & 77.34  \\

    \hline
    
    \end{tabular}
  \end{minipage} 
\end{table}

\subsection{Action Recognition}
\label{sec:act}
\textbf{Baseline Methods.} For the action recognition task, we adopt three baselines to evaluate the effectiveness of our CGC: TSN~\cite{TSN}, P3D-A~\cite{qiu2017learning} (details are in the supplementary material), and TSM~\cite{lin2018temporal}. Because our CGC's effectiveness of introducing 2D spatial context to CNNs has been verified on image classification, in this part, we focus on its ability of incorporating 1D temporal context and 3D spatiotemporal context. For the 1D case, we apply our CGC to temporal convolutions in every P3D-A block. For the 3D case, we apply our CGC to spatial convolutions in P3D-A or 2D convolutions in TSN or TSM; the pooling layer produces $c\times k\times k \times k$ cubes, the Context Encoding Module encodes $k\times k \times k$ feature maps into a vector of length $k^3/2$, and the Gate Decoding Module generates $o\times c \times t\times k \times k$ gates. Note that for the first convolutional layer, we use $\tI \in \R^{3\times 64}$ for the Channel Interacting Module.

\begin{table*}[t]
\caption{Action recognition results on Something-Something (v1). Backbone indicates the backbone network architecture. Param indicates the number of parameters in the model. Frame indicates the number of frames used for evaluation. Bold indicates the best result.}
\label{tab:sth}
\begin{center}
\tiny
\begin{tabular}{llllll}
\hline
Model  & Backbone & Param  & Frame & Top-1(\%) & Top-5(\%) 
\\ \hline 
TRN~\cite{zhou2018temporal} & BNInception         & 18.3M & 8 & 34.4 & - \\ 
TRN~\cite{lin2018temporal} & ResNet-50          & 31.8M & 8 & 38.9 & 68.1 \\ 
ECO~\cite{zolfaghari2018eco} & BNInc+Res18  & 47.5M & 8 & 39.6 &- \\
ECO~\cite{zolfaghari2018eco} & BNInc+Res18 & 47.5M & 16 & 41.4 &- \\
ECO$_{En}$Lite~\cite{zolfaghari2018eco} & BNInc+Res18 & 150M & 92 & 46.4 &- \\
\hline \hline
TSN~\cite{TSN} & ResNet-50         & 23.86M & 8 & 19.00 & 44.98 \\ 
TSN + Non-local~\cite{wang2017non} & ResNet-50         & 31.22M & 8 & 25.73 & 55.17 \\ 
\textbf{TSN + CGC (Ours)} & ResNet-50         & 24.07M & 8 & \textbf{32.58} &  \textbf{60.06} \\
\hline
P3D~\cite{qiu2017learning} & ResNet-50         & 25.38M & $32\times 30$ & 45.17 & 74.61 \\ 
lslsP3D + Non-local~\cite{wang2017non} & ResNet-50         & 32.73M & $32\times 30$ & 45.88 & 74.94 \\ 
\textbf{P3D + CGC 1D (Ours)} & ResNet-50         & 25.39M & $32\times 30$ & 46.14 & 75.92 \\
\textbf{P3D + CGC 3D (Ours)} & ResNet-50         & 25.61M & $32\times 30$ & 46.35 &  75.97 \\
\textbf{P3D + CGC 1D \& 3D (Ours)} & ResNet-50         & 25.62M & $32\times 30$ & \textbf{46.73} &  \textbf{76.04} \\
\hline
TSM~\cite{lin2018temporal} & ResNet-50         & 23.86M & 8 & 44.65 & 73.94 \\ 
TSM + Non-local~\cite{wang2017non} & ResNet-50         & 31.22M & 8 & 43.91 & 72.18 \\ 
\textbf{TSM + CGC (Ours)} & ResNet-50         & 24.07M & 8 & \textbf{46.00} &  \textbf{75.11} \\
\hline
TSM~\cite{lin2018temporal} & ResNet-50         & 23.86M & 16 & 46.61 & 76.18 \\ 
\textbf{TSM + CGC (Ours)} & ResNet-50         & 24.09M & 16 & \textbf{47.87} &  \textbf{77.22} \\
\hline
\end{tabular}
\end{center}

\end{table*}

\begin{table*}[t]
\caption{Action recognition results on Kinetics. Backbone indicates the backbone network architecture. Param indicates the number of parameters in the model. Bold indicates the best result.}
\label{tab:kin}
\begin{center}
\tiny
\begin{tabular}{lllll}
\hline

Model  & Backbone & Param   & Top-1(\%) & Top-5(\%) 
\\ \hline

TSM~\cite{lin2018temporal} & ResNet-50         & 23.86M  & 74.12 & 91.21 \\ 
TSM + Non-local~\cite{wang2017non} & ResNet-50         & 31.22M  & 75.60 & 92.15 \\ 
\textbf{TSM + CGC (Ours)} & ResNet-50         & 24.07M  & 76.06 &  \textbf{92.50} \\
\textbf{TSM + Non-local + CGC (Ours)} & ResNet-50         & 31.43M & \textbf{76.40} &  \textbf{92.50} \\
\hline

\end{tabular}
\end{center}

\end{table*}

\textbf{Experimental Setting.} 
The Something-Something (v1) dataset has a training split of 86,017 videos and a validation split of 
11,522 videos, with 174 categories. We follow~\cite{qiao2019weight} to train on the training set and report evaluation results on the validation set. We follow~\cite{lin2018temporal} to process videos and augment data. Since we only use ImageNet for pretraining, we adapt the code base of TSM but the training setting from~\cite{qiao2019weight}. We train TSN- and TSM-based models for 45 epochs (50 for P3D-A), starting from a learning rate of 0.025 (0.01 for P3D-A) and decreasing it by 0.1 at 26 and 36 epochs (30, 40, 45 for P3D-A). The Kinetics~\cite{Kinetics} dataset has 400 action classes and 240K training samples. We follow~\cite{lin2018temporal} to train and evaluate all the compared models.

For TSN- and TSM-based models, the batch size is 64 for 8-frame models and 32 for 16-frame models, and the dropout rate is set to 0.5. P3D-A takes 32 continuously sampled frames as input and the batch size is 64, and the dropout ratio is 0.8. We use the evaluation setting of~\cite{lin2018temporal} for TSN- and TSM-based models and the evaluation settings of~\cite{wang2017non} for P3D-A. All the models are trained with 8-GPU machines.

\textbf{Performance Comparisons.} 
As Table~\ref{tab:sth} and Table~\ref{tab:kin} show, our CGC significantly improves the performance of baseline CNN models, compared to Non-local~\cite{wang2017non}. As aforementioned, Non-local modules modify the input feature maps of convolutional layers by reassembling local features according to the global correspondence. We apply Non-local blocks in the most effective way as is reported by~\cite{wang2017non}. However, we observe that its performance gain is not consistent when training the model from scratch. When applied to TSM on the Something-Something dataset, it even degrades the performance. Our proposed CGC consistently improves the performances of all the baseline models. We also observe that on Kinetics, our CGC and Non-local are somewhat complementary to each other since applying both of them to the baseline achieves the highest performance. This is consistent with the observation of the combination of CBAM and our CGC in Section~\ref{sec:image}.

\subsection{Machine Translation}
\label{exp:lab}
\textbf{Baseline Methods.} The LightConv proposed by~\cite{wu2019pay} achieves better performance with a lightweight convolutional model, compared to Transformer~\cite{vaswani2017attention}. We take it as the baseline model and augment its Lightweight Convolution with our CGC. Note that the Lightweight Convolution is a grouped convolution $\tL \in \R^{H \times k}$ with weight sharing, so we remove the Channel Interacting Module since we do not need it to project latent representations. We resize the input sequence $\mS \in \R ^{c \times L}$ to $\R ^{H \times 3k}$ with average pooling. For those sequences shorter than $3k$, we pad them with zeros. Since the decoder decodes translated words one by one at the inference time, it is unclear how to define global context for it. Therefore, we only replace the convolutions in the encoder.

\textbf{Experimental Setting.} 
We follow~\cite{wu2019pay} to train all the compared models with 160K sentence pairs and 10K joint BPE vocabulary. We use the training protocol of DynamicConv~\cite{wu2019pay} provided in~\cite{ott2019fairseq}. The widely-used BLEU-4~\cite{papineni2002bleu} is reported for evaluation of all the models. We find that it is necessary to set beam width to 6 to reproduce the results of DynamicConv reported in~\cite{wu2019pay}, and we fix it to be 6 for all the models.

\begin{table}[t]
\caption{Machine translation results on IWSLT'14 De-En. Param indicates the number of parameters in the model. Bold indicates the best result.}
\label{tab:mt}
\scriptsize
\begin{center}

\begin{tabular}{lll}
\hline
Model  & Param & {BLEU-4} 
\\ \hline 
Deng \textit{et al.}~\cite{deng2018latent} & - & 33.08 \\
Transformer~\cite{vaswani2017attention} & 39.47M & 34.41 \\
\hline
LightConv~\cite{wu2019pay} & 38.14M & 34.84 \\
LightConv + Dynamic Encoder~\cite{wu2019pay} & 38.44M & 35.03 \\
\textbf{LightConv + CGC Encoder (Ours)} & \textbf{38.15M} & \textbf{35.21} \\
\hline
\end{tabular}
\end{center}

\end{table}

\textbf{Performance Comparisons.}
As Table~\ref{tab:mt} shows, replacing Lightweight Convolutions in the encoder of LightConv with our CGC significantly outperforms LightConv and LightConv + Dynamic Encoder by 0.37 and 0.18 BLEU, respectively, yielding the state-of-the-art performance. As was discussed previously, Dynamic Convolution leverages a linear layer to generate the convolution kernel according to the input segment, which lacks the awareness of global context. This flaw may lead to sub-optimal encoding of the source sentence and thus the unsatisfying decoded sentence. However, our CGC incorporates global context of the source sentence and helps significantly improve the quality of the translated sentence. Moreover, our CGC is much more efficient than Dynamic Convolution because of our module design. Our CGC only needs 0.01M extra parameters, but Dynamic Convolution needs $30\times$ more.

\section{Conclusions}
In this paper, motivated by the neuroscience research on neurons as ``\textit{adaptive processors}'', we proposed a lightweight Context-Gated Convolution (CGC) to incorporate global context information into CNNs. Different from previous works which usually modify input feature maps, our proposed CGC directly modulates convolution kernels under the guidance of global context information. In specific, we proposed three modules to efficiently generate a gate to modify the kernel. As such, our CGC is able to extract representative local patterns according to global context. The extensive experimental results show consistent performance improvements on various tasks with a negligible computational complexity and parameter increment. In the future, our proposed CGC can be incorporated into the searching space of Neural Architecture Search (NAS) to further improve the performance of NAS models.
%
%
\bibliographystyle{splncs04}
\bibliography{egbib}

\begin{thebibliography}{10}
\providecommand{\url}[1]{\texttt{#1}}
\providecommand{\urlprefix}{URL }
\providecommand{\doi}[1]{https://doi.org/#1}

\bibitem{ba2016layer}
Ba, J.L., Kiros, J.R., Hinton, G.E.: Layer normalization. arXiv preprint
  arXiv:1607.06450  (2016)

\bibitem{barbu2019objectnet}
Barbu, A., Mayo, D., Alverio, J., Luo, W., Wang, C., Gutfreund, D., Tenenbaum,
  J., Katz, B.: Objectnet: A large-scale bias-controlled dataset for pushing
  the limits of object recognition models. In: Advances in Neural Information
  Processing Systems. pp. 9448--9458 (2019)

\bibitem{bello2019attention}
Bello, I., Zoph, B., Vaswani, A., Shlens, J., Le, Q.V.: Attention augmented
  convolutional networks. arXiv preprint arXiv:1904.09925  (2019)

\bibitem{cao2019gcnet}
Cao, Y., Xu, J., Lin, S., Wei, F., Hu, H.: Gcnet: Non-local networks meet
  squeeze-excitation networks and beyond. arXiv preprint arXiv:1904.11492
  (2019)

\bibitem{Kinetics}
Carreira, J., Zisserman, A.: Quo vadis, action recognition? a new model and the
  kinetics dataset. In: CVPR (2017)

\bibitem{cettolo2014report}
Cettolo, M., Niehues, J., St{\"u}ker, S., Bentivogli, L., Federico, M.: Report
  on the 11th iwslt evaluation campaign, iwslt 2014

\bibitem{chattopadhay2018grad}
Chattopadhay, A., Sarkar, A., Howlader, P., Balasubramanian, V.N.: Grad-cam++:
  Generalized gradient-based visual explanations for deep convolutional
  networks. In: 2018 IEEE Winter Conference on Applications of Computer Vision
  (WACV). pp. 839--847. IEEE (2018)

\bibitem{chen2019graph}
Chen, Y., Rohrbach, M., Yan, Z., Shuicheng, Y., Feng, J., Kalantidis, Y.:
  Graph-based global reasoning networks. In: Proceedings of the IEEE Conference
  on Computer Vision and Pattern Recognition. pp. 433--442 (2019)

\bibitem{cheng2018dual}
Cheng, C., Fu, Y., Jiang, Y.G., Liu, W., Lu, W., Feng, J., Xue, X.: Dual
  skipping networks. In: Proceedings of the IEEE Conference on Computer Vision
  and Pattern Recognition. pp. 4071--4079 (2018)

\bibitem{child2019generating}
Child, R., Gray, S., Radford, A., Sutskever, I.: Generating long sequences with
  sparse transformers. arXiv preprint arXiv:1904.10509  (2019)

\bibitem{chollet2017xception}
Chollet, F.: Xception: Deep learning with depthwise separable convolutions. In:
  Proceedings of the IEEE conference on computer vision and pattern
  recognition. pp. 1251--1258 (2017)

\bibitem{dai2017deformable}
Dai, J., Qi, H., Xiong, Y., Li, Y., Zhang, G., Hu, H., Wei, Y.: Deformable
  convolutional networks. In: Proceedings of the IEEE international conference
  on computer vision. pp. 764--773 (2017)

\bibitem{deng2018latent}
Deng, Y., Kim, Y., Chiu, J., Guo, D., Rush, A.: Latent alignment and
  variational attention. In: Advances in Neural Information Processing Systems.
  pp. 9712--9724 (2018)

\bibitem{gehring2017convolutional}
Gehring, J., Auli, M., Grangier, D., Yarats, D., Dauphin, Y.N.: Convolutional
  sequence to sequence learning. In: Proceedings of the 34th International
  Conference on Machine Learning-Volume 70. pp. 1243--1252. JMLR. org (2017)

\bibitem{gilbert2013top}
Gilbert, C.D., Li, W.: Top-down influences on visual processing. Nature Reviews
  Neuroscience  \textbf{14}(5), ~350 (2013)

\bibitem{rcnn}
Girshick, R., Donahue, J., Darrell, T., Malik, J.: Rich feature hierarchies for
  accurate object detection and semantic segmentation. In: CVPR (2014)

\bibitem{pmlr-v9-glorot10a}
Glorot, X., Bengio, Y.: Understanding the difficulty of training deep
  feedforward neural networks. In: Teh, Y.W., Titterington, M. (eds.)
  Proceedings of the Thirteenth International Conference on Artificial
  Intelligence and Statistics. Proceedings of Machine Learning Research,
  vol.~9, pp. 249--256. PMLR, Chia Laguna Resort, Sardinia, Italy (13--15 May
  2010), \url{http://proceedings.mlr.press/v9/glorot10a.html}

\bibitem{Goyal_2017_ICCV}
Goyal, R., Ebrahimi~Kahou, S., Michalski, V., Materzynska, J., Westphal, S.,
  Kim, H., Haenel, V., Fruend, I., Yianilos, P., Mueller-Freitag, M., Hoppe,
  F., Thurau, C., Bax, I., Memisevic, R.: The "something something" video
  database for learning and evaluating visual common sense. In: The IEEE
  International Conference on Computer Vision (ICCV) (Oct 2017)

\bibitem{ha2016hypernetworks}
Ha, D., Dai, A., Le, Q.V.: Hypernetworks. arXiv preprint arXiv:1609.09106
  (2016)

\bibitem{He_2015}
He, K., Zhang, X., Ren, S., Sun, J.: Delving deep into rectifiers: Surpassing
  human-level performance on imagenet classification. 2015 IEEE International
  Conference on Computer Vision (ICCV)  (Dec 2015).
  \doi{10.1109/iccv.2015.123}, \url{http://dx.doi.org/10.1109/ICCV.2015.123}

\bibitem{He_2016_CVPR}
He, K., Zhang, X., Ren, S., Sun, J.: Deep residual learning for image
  recognition. In: CVPR (2016)

\bibitem{he2016identity}
He, K., Zhang, X., Ren, S., Sun, J.: Identity mappings in deep residual
  networks. In: European conference on computer vision. pp. 630--645. Springer
  (2016)

\bibitem{he2019bag}
He, T., Zhang, Z., Zhang, H., Zhang, Z., Xie, J., Li, M.: Bag of tricks for
  image classification with convolutional neural networks. In: Proceedings of
  the IEEE Conference on Computer Vision and Pattern Recognition. pp. 558--567
  (2019)

\bibitem{howard2017mobilenets}
Howard, A.G., Zhu, M., Chen, B., Kalenichenko, D., Wang, W., Weyand, T.,
  Andreetto, M., Adam, H.: Mobilenets: Efficient convolutional neural networks
  for mobile vision applications. arXiv preprint arXiv:1704.04861  (2017)

\bibitem{hu2018squeeze}
Hu, J., Shen, L., Sun, G.: Squeeze-and-excitation networks. In: Proceedings of
  the IEEE conference on computer vision and pattern recognition. pp.
  7132--7141 (2018)

\bibitem{huang2017densely}
Huang, G., Liu, Z., Van Der~Maaten, L., Weinberger, K.Q.: Densely connected
  convolutional networks. In: CVPR (2017)

\bibitem{ioffe2015batch}
Ioffe, S., Szegedy, C.: Batch normalization: Accelerating deep network training
  by reducing internal covariate shift. In: ICML (2015)

\bibitem{jia2016dynamic}
Jia, X., De~Brabandere, B., Tuytelaars, T., Gool, L.V.: Dynamic filter
  networks. In: Advances in Neural Information Processing Systems. pp. 667--675
  (2016)

\bibitem{jo2018deep}
Jo, Y., Wug~Oh, S., Kang, J., Joo~Kim, S.: Deep video super-resolution network
  using dynamic upsampling filters without explicit motion compensation. In:
  Proceedings of the IEEE Conference on Computer Vision and Pattern
  Recognition. pp. 3224--3232 (2018)

\bibitem{kim2014convolutional}
Kim, Y.: Convolutional neural networks for sentence classification. arXiv
  preprint arXiv:1408.5882  (2014)

\bibitem{krizhevsky2009learning}
Krizhevsky, A., et~al.: Learning multiple layers of features from tiny images.
  Tech. rep., Citeseer (2009)

\bibitem{li2004perceptual}
Li, W., Pi{\"e}ch, V., Gilbert, C.D.: Perceptual learning and top-down
  influences in primary visual cortex. Nature neuroscience  \textbf{7}(6),
  651--657 (2004)

\bibitem{li2019selective}
Li, X., Wang, W., Hu, X., Yang, J.: Selective kernel networks. In: Proceedings
  of the IEEE Conference on Computer Vision and Pattern Recognition. pp.
  510--519 (2019)

\bibitem{lin2018temporal}
Lin, J., Gan, C., Han, S.: Temporal shift module for efficient video
  understanding. arXiv preprint arXiv:1811.08383  (2018)

\bibitem{mildenhall2018burst}
Mildenhall, B., Barron, J.T., Chen, J., Sharlet, D., Ng, R., Carroll, R.: Burst
  denoising with kernel prediction networks. In: Proceedings of the IEEE
  Conference on Computer Vision and Pattern Recognition. pp. 2502--2510 (2018)

\bibitem{nair2010rectified}
Nair, V., Hinton, G.E.: Rectified linear units improve restricted boltzmann
  machines. In: Proceedings of the 27th international conference on machine
  learning (ICML-10). pp. 807--814 (2010)

\bibitem{ott2019fairseq}
Ott, M., Edunov, S., Baevski, A., Fan, A., Gross, S., Ng, N., Grangier, D.,
  Auli, M.: fairseq: A fast, extensible toolkit for sequence modeling. In:
  Proceedings of NAACL-HLT 2019: Demonstrations (2019)

\bibitem{papineni2002bleu}
Papineni, K., Roukos, S., Ward, T., Zhu, W.J.: Bleu: a method for automatic
  evaluation of machine translation. In: Proceedings of the 40th annual meeting
  on association for computational linguistics. pp. 311--318. Association for
  Computational Linguistics (2002)

\bibitem{park2018bam}
Park, J., Woo, S., Lee, J.Y., Kweon, I.S.: Bam: Bottleneck attention module.
  arXiv preprint arXiv:1807.06514  (2018)

\bibitem{parmar2018image}
Parmar, N., Vaswani, A., Uszkoreit, J., Kaiser, {\L}., Shazeer, N., Ku, A.,
  Tran, D.: Image transformer. arXiv preprint arXiv:1802.05751  (2018)

\bibitem{pytorch}
Paszke, A., Gross, S., Chintala, S., Chanan, G., Yang, E., DeVito, Z., Lin, Z.,
  Desmaison, A., Antiga, L., Lerer, A.: Automatic differentiation in pytorch
  (2017)

\bibitem{qiao2019weight}
Qiao, S., Wang, H., Liu, C., Shen, W., Yuille, A.: Weight standardization.
  arXiv preprint arXiv:1903.10520  (2019)

\bibitem{qiu2017learning}
Qiu, Z., Yao, T., Mei, T.: Learning spatio-temporal representation with
  pseudo-3d residual networks. In: 2017 IEEE International Conference on
  Computer Vision (ICCV). pp. 5534--5542. IEEE (2017)

\bibitem{fasterrcnn}
Ren, S., He, K., Girshick, R., Sun, J.: Faster r-cnn: Towards real-time object
  detection with region proposal networks. In: NIPS (2015)

\bibitem{russakovsky2015imagenet}
Russakovsky, O., Deng, J., Su, H., Krause, J., Satheesh, S., Ma, S., Huang, Z.,
  Karpathy, A., Khosla, A., Bernstein, M., et~al.: Imagenet large scale visual
  recognition challenge. International journal of computer vision
  \textbf{115}(3),  211--252 (2015)

\bibitem{santurkar2018does}
Santurkar, S., Tsipras, D., Ilyas, A., Madry, A.: How does batch normalization
  help optimization? In: Advances in Neural Information Processing Systems. pp.
  2483--2493 (2018)

\bibitem{stollenga2014deep}
Stollenga, M.F., Masci, J., Gomez, F., Schmidhuber, J.: Deep networks with
  internal selective attention through feedback connections. In: Advances in
  neural information processing systems. pp. 3545--3553 (2014)

\bibitem{vaswani2017attention}
Vaswani, A., Shazeer, N., Parmar, N., Uszkoreit, J., Jones, L., Gomez, A.N.,
  Kaiser, {\L}., Polosukhin, I.: Attention is all you need. In: Advances in
  neural information processing systems. pp. 5998--6008 (2017)

\bibitem{wang2017residual}
Wang, F., Jiang, M., Qian, C., Yang, S., Li, C., Zhang, H., Wang, X., Tang, X.:
  Residual attention network for image classification. In: Proceedings of the
  IEEE Conference on Computer Vision and Pattern Recognition. pp. 3156--3164
  (2017)

\bibitem{TSN}
Wang, L., Xiong, Y., Wang, Z., Qiao, Y., Lin, D., Tang, X., Gool, L.V.:
  Temporal segment networks: Towards good practices for deep action
  recognition. In: ECCV (2016)

\bibitem{wang2017non}
Wang, X., Girshick, R., Gupta, A., He, K.: Non-local neural networks. arXiv
  preprint arXiv:1711.07971  \textbf{10} (2017)

\bibitem{woo2018cbam}
Woo, S., Park, J., Lee, J.Y., So~Kweon, I.: Cbam: Convolutional block attention
  module. In: Proceedings of the European Conference on Computer Vision (ECCV).
  pp. 3--19 (2018)

\bibitem{wu2019pay}
Wu, F., Fan, A., Baevski, A., Dauphin, Y.N., Auli, M.: Pay less attention with
  lightweight and dynamic convolutions. arXiv preprint arXiv:1901.10430  (2019)

\bibitem{yang2018convolutional}
Yang, Y., Zhong, Z., Shen, T., Lin, Z.: Convolutional neural networks with
  alternately updated clique. In: Proceedings of the IEEE Conference on
  Computer Vision and Pattern Recognition. pp. 2413--2422 (2018)

\bibitem{zamir2017feedback}
Zamir, A.R., Wu, T.L., Sun, L., Shen, W.B., Shi, B.E., Malik, J., Savarese, S.:
  Feedback networks. In: Proceedings of the IEEE Conference on Computer Vision
  and Pattern Recognition. pp. 1308--1317 (2017)

\bibitem{zhang2017mixup}
Zhang, H., Cisse, M., Dauphin, Y.N., Lopez-Paz, D.: mixup: Beyond empirical
  risk minimization. arXiv preprint arXiv:1710.09412  (2017)

\bibitem{zhang2015character}
Zhang, X., Zhao, J., LeCun, Y.: Character-level convolutional networks for text
  classification. In: Advances in neural information processing systems. pp.
  649--657 (2015)

\bibitem{zhou2018temporal}
Zhou, B., Andonian, A., Oliva, A., Torralba, A.: Temporal relational reasoning
  in videos. In: Proceedings of the European Conference on Computer Vision
  (ECCV). pp. 803--818 (2018)

\bibitem{CycleGAN2017}
Zhu, J.Y., Park, T., Isola, P., Efros, A.A.: Unpaired image-to-image
  translation using cycle-consistent adversarial networks. In: Computer Vision
  (ICCV), 2017 IEEE International Conference on (2017)

\bibitem{zhu2019deformable}
Zhu, X., Hu, H., Lin, S., Dai, J.: Deformable convnets v2: More deformable,
  better results. In: Proceedings of the IEEE Conference on Computer Vision and
  Pattern Recognition. pp. 9308--9316 (2019)

\bibitem{zolfaghari2018eco}
Zolfaghari, M., Singh, K., Brox, T.: Eco: Efficient convolutional network for
  online video understanding. In: Proceedings of the European Conference on
  Computer Vision (ECCV). pp. 695--712 (2018)

\end{thebibliography}

\appendix
\section{Appendix}

\subsection{Analysis of the Gate}
\begin{figure*}[t]

\begin{center}
\includegraphics[width=0.99\linewidth]{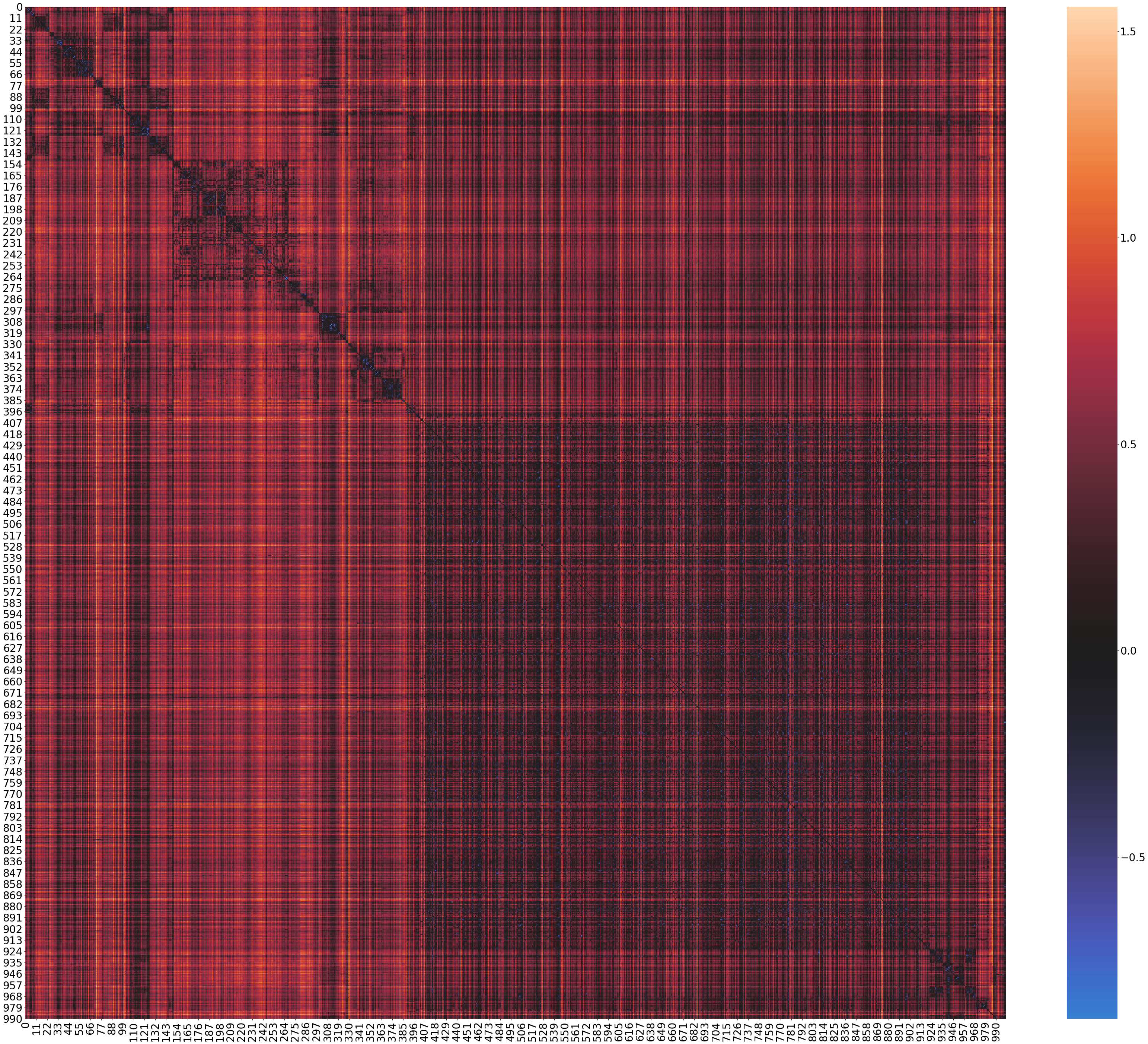}
\end{center}
\caption{Visualization of the difference matrix between inter-class distances and intra-class distances of the last gate in the network on the ImageNet validation set. (Best viewed on a monitor when zoomed in)}
\label{dist}
\end{figure*}

To further validate that our CGC uses context information of the target objects to guide the convolution process, we calculate the average modulated kernel (in the last CGC of the model) for images of each class in the validation set. Then we calculate inter-class $L-2$ distance between every two average modulated kernels, i.e., class centers, and the intra-class $L-2$ distance (mean distance to the class center) for each class. As is shown in Fig.~\ref{dist}, we visualize the difference matrix between inter-class distances and intra-class distances. In more than 93.99\% of the cases, the inter-class distance is larger than the corresponding intra-class distance, which indicates that there are clear clusters of these modulated kernels and the clusters are aligned very well with the classes.

This observation strongly supports that our CGC successfully extracts class-specific context information and effectively modulates the convolution kernel to extract representative features. Meanwhile, the intra-class variance of the modulated kernels indicates that our CGC dynamically modulates convolution kernels according to different input contexts.

\subsection{Details of training settings on ImageNet and CIFAR-10}
\label{setting}
For the default setting on ImageNet, we use $224\times 224$ random resized cropping and random horizontal flipping for data augmentation. Then we standardize the data with mean and variance per channel. We use a traditional cross-entropy loss to train all the networks with a batch size of 256 on 8 GPUs by SGD, a weight decay of 0.0001, and a momentum of 0.9 for 100 epochs. We start from a learning rate of 0.1 and decrease it by a factor of 10 every 30 epochs. The last normalization layers in the module are zero-initialized to make the gates start from constants. All the extra layers in Context-Gated Convolution have a learning rate ten times smaller than convolutional kernels.

For the advanced setting, we also use mixup~\cite{zhang2017mixup} for data augmentation, and we follow~\cite{he2019bag} to use learning rate warm-up in the first 5 epochs of training. We train the networks with the cosine learning rate schedule~\cite{he2019bag} for 120 epochs. The other hyper-parameters are set to be same with the default setting.

For CIFAR-10, we use $32\times 32$ random cropping with a padding of 4 and random horizontal flipping. We use a batch size of 128 and train on 1 GPU. We decrease the learning rate at the 81st and 122nd epochs, and halt training after 164 epochs. 
For the ablation study, the result is an average of 3 runs.

\subsection{Details about P3D-A}
Based on ResNet-50, we add a temporal convolution with $k=5,stride=2$ after the first convolutional layer. For convolutional layers in residual blocks, we follow~\cite{wang2017non} to add $3\times 1 \times 1$ convolution (stride is 1) after every two $1\times 3 \times 3$ convolutions. The added temporal convolutional layers are initialized to imitate the behavior of TSM~\cite{lin2018temporal} to ease the training process. We only inflate the max pooling layer after the first convolutional layer with a temporal kernel size of 3 and a stride of 2 without adding any other temporal pooling layers. Note that all the aforementioned convolutional layers come with a Batch Normalization layer and a ReLU activation function.

\end{document}